\documentclass[runningheads]{llncs}
\usepackage{graphicx}
\usepackage{gb4e}
\usepackage{multirow}
\usepackage{array}
\usepackage{soul}
\usepackage{academicons}

\begin{document}
\title{Labeling Explicit Discourse Relations using Pre-trained Language Models}

%
\author{Murathan Kurfal{\i}\orcidID{0000-0002-7020-8275}}

\authorrunning{M. Kurfal{\i}}
%
\institute{Linguistics Department, Stockholm University, Stockholm, Sweden}
\maketitle
\begin{abstract}
Labeling explicit discourse relations is one of the most challenging sub-tasks of the shallow discourse parsing where the goal is to identify the discourse connectives and the boundaries of their arguments. The state-of-the-art models achieve slightly above 45\% of F-score by using hand-crafted features. The current paper investigates the efficacy of the pre-trained language models in this task. We find that the pre-trained language models, when finetuned, are powerful enough to replace the linguistic features. We evaluate our model on PDTB 2.0 and report the state-of-the-art results in extraction of the full relation. This is the first time when a model outperforms the knowledge intensive models without employing any linguistic features. 

\end{abstract}
%
%
%


\section{Introduction}

Shallow discourse parsing (SDP) refers to the task of segmenting a text into a set of discourse relations. A typical discourse relation consists of two arguments and a discourse connective accompanied with a sense reflecting the semantic relation between the arguments (e.g. cause, precedence). Within the Penn Discourse Treebank (PDTB), discourse connectives are assumed to be the lexical items which connect two abstract objects such as events, states, and propositions following the definition of \cite{prasad2008penn}. There are two main types of discourse relations, \textit{explicit} and \textit{implicit}, where the difference is the presence of an overt discourse connective. Parsing explicit and implicit relations are often treated as separate tasks, and implicit discourse relations have received much of the attention due to the challenges brought by a lack of an overt signal. In this work, we instead focus on the less studied task of identifying explicit discourse relations. This consists of identifying discourse connectives and their arguments in text.

Labeling explicit relations is a challenging task due to three main reasons: (i) connectives do not always assume a discursive role (ii) the arguments can consist of discontinuous text spans (iii) the same text span can have different roles in different relations. All three challenges are illustrated in the Example \ref{ex:also} and \ref{ex:alt}\footnote{In the examples, unless otherwise stated, Arg2 is
shown in bold, Arg1 is in italics and the discourse connective is underlined}.

\begin{exe}
    \ex  \textit{Although Dr. Warshaw points out that stress and anxiety have their positive uses, \lq stress perceived to be threatening implies a component of fear and anxiety that may contribute to burnout.\rq} \textbf{He} \underline{also} \textbf{noted that various work environments, such as night work, have their own stressors}. \label{ex:also}
    
    \ex  \underline{Although} Dr. Warshaw points out that\textbf{ stress and anxiety have their positive uses}, \textit{"stress perceived to be threatening implies a component of fear and anxiety that may contribute to burnout"} \label{ex:alt}

\end{exe}

Example \ref{ex:also} presents a case where a complete discourse relation, which is provided in Example \ref{ex:alt}, is embedded within another relation. Therefore, the text span \lq stress and ... uses\rq\ assumes two different roles in two different relations; it is part of the first argument in the first relation, whereas it is the second argument in Example \ref{ex:alt}.  Additionally, the second argument in Example \ref{ex:also} consists of an discontinuous text span as it is is interrupted by the connective. Similarly, the text span (\lq Dr. Warshaw ... that\rq) creates discontinuity between the connective and the second argument in Example \ref{ex:alt} as it does not belong to the relation at all. Finally, the lexical item \textit{and}, which is the most common discourse connective, in Example \ref{ex:alt} do not assume any discursive role in this case as it only connects two noun phrases rather than abstract objects. 

Most existing literature heavily relies on feature engineering to deal with these issues, with \cite{hooda2017argument} and \cite{knaebel2019window} being the only notable exceptions. The current work follows the latter studies in performing explicit discourse relation labeling without resorting to any linguistic features. Specifically, we try to answer the following question: can pre-trained language models, which have shown significant gains in a wide variety of natural language tasks, replace the rich feature sets used in a standard parser? To this end, we model explicit discourse relation labeling as a pipeline of two tasks, namely connective identification and argument labeling. Each sub-task is regarded as a token-level sequence prediction problem and modeled as a simple neural architecture employing BERT \cite{devlin2018bert}. 

We evaluated our system on the PDTB 2.0 corpus. Experimental results show that contextual embeddings are indeed powerful enough to replace the linguistics features used in previous work. Unlike the previous feature-independent models, the proposed system manages to improve over the existing systems by achieving 8\% increase in the extraction of the both arguments\footnote{The source code will be available at: https://github.com/MurathanKurfali/shallowdisc}. Besides the performance gain, the proposed system has the benefit of being directly applicable to any raw text as it does not require any pre-processing and can straightforwardly be extended to other languages with available training data and a pre-trained language model.

\section{Background}

Shallow discourse parsing (SDP) aims to uncover the local coherence relations within text without assuming any tree/graph structure between the relations, hence the name \textit{shallow}. It started with the release of PDTB 2.0 \cite{prasad2008penn} and, lately, it attracted attention thanks to the two subsequent shared tasks at CoNLL 2015 \cite{xue2015conll} and 2016 \cite{xue2016conll}. Most of the participating systems treat SDP as a pipeline of sub-tasks following \cite{lin2014pdtb} which is the first full end-to-end PDTB-style discourse parser. A standard pipeline starts with labeling explicit discourse relations which is followed by sense classification and labeling Implicit discourse relations. Labeling explicit relations are further decomposed into a set of sub-tasks which are connective identification, argument position identification, extraction of each argument. Each sub-task is addressed by employing  a rich set of linguistics features, including dependency trees, part-of-speech tags, Brown clusters and Verbnet classes \cite{xue2015conll,xue2016conll}. 

\cite{hooda2017argument} marks the beginning of a new line of research which is to perform shallow discourse parsing without any feature engineering. The authors address labeling of explicit discourse relations task in a simplified setting where the task is reduced to determining the role of each token within a pre-extracted relation span. The authors train a LSTM on those spans which takes Glove embeddings as its input and classifies each token with one of the four labels which are \textit{Conn, Arg1, Arg2, None}. The network achieves F-score of 23.05\% which is significantly lower than the state-of-the-art models. Nevertheless, the study is of great importance as it shows that argument labeling is possible without any feature engineering. \cite{knaebel2019window} extends the idea of \cite{hooda2017argument} to full shallow discourse parsing on raw texts. They employ a BiLSTM and a sliding window approach, according to which the text is split into overlapping windows. In each window the system tries to capture the parts which belong to a discourse relation. The predicted relation spans are later assembled using a novel aggregation method based on Jaccard distance. The proposed hierarchy performs considerably better than \cite{hooda2017argument} but still falls short of matching the state-of-the-art methods.

Method-wise, the closest work to the current paper is that of \cite{muller2019tony} (ToNy) which employs contextual embeddings to perform multilingual RST-style discourse segmentation. Instead of directly finetuning BERT, ToNy uses a simplified sequence prediction architecture which consists of an LSTM conditioned on the concatenation of the contextual embeddings (either Elmo or BERT) and the character embeddings obtained by convolution filters.

\section{Method}

Unlike previous studies which realize labeling explicit discourse relations as a long pipeline of tasks which usually consists of 4 to 6 sub-components \cite{wang2015refined,qin2016shallow}, we propose a simplified pipeline consisting of only two steps, namely connective identification and argument extraction. 

The connective identification step helps us to exploit the lexicalized approach of the PDTB. In the PDTB framework, the discourse connectives function as anchor points and without them, determining the number of relations in a text become highly problematic, especially when the arguments of multiple relations overlap as in Example \ref{ex:also} and \ref{ex:alt}. Bypassing connective identification  would require an extra post-processing step to sort different relations with common arguments out which is not a trivial task. In order to avoid those problem, we perform connective identification as the first task, mimicking the original annotation schema. 

Following the previous studies of \cite{hooda2017argument,knaebel2019window}, we approach explicit discourse relation labeling as an N-way token classification problem. To this end, we follow the standard token classification architecture used in sequence prediction tasks, e.g. named entity recognition, employing BERT \cite{devlin2018bert}. The architecture consists of a BERT model with a linear layer on top. The linear layer is connected to the hidden-states of the BERT and outputs the label probabilities for each token based on the sub-task it is trained on.

\subsection{Connective Identification} \label{sec:conn}
The aim of this component is to identify the lexical items which assume a discursive role in the text. Although connective identification seems to be the easiest step among the other sub-tasks of shallow discourse parsing \cite{li2016constituent,qin2016shallow}, it has its own challenges. One problem is that discourse connectives in PDTB can be multiword expressions such as lexically frozen phrases, e.g. \textit{on the other hand}, or as modified connectives which co-occur with an adverb e.g. \textit{partly because, particularly since}. Such multi-word connectives pose a challenge because different connectives may also appear in the text consecutively without forming a longer connective, as illustrated in Example \ref{ex:conn1} and \ref{ex:mweconn}. 
\begin{exe} 
	\ex   \label{ex:conn1}
	\begin{xlist} 
	\ex \textit{Typically, developers option property}, \underline{and} \textbf{then once they get the administrative approvals, they buy it} (Conjunction)
	\ex \textit{Typically, developers option property}, \textbf{and} \underline{then} \textbf{once they get the administrative approvals, they buy it} (Precedence) 
	\ex  Typically, developers option property, \textit{and then}  \underline{once} \textbf{they get the administrative approvals}, \textit{they buy it}   (Succession)   (WSJ\_2313)
	\end{xlist} 
	\ex   \textit{Consider the experience of Satoko Kitada, a 30-year-old designer of vehicle interiors who joined Nissan in 1982}. \underline{At that time}, \textbf{tasks were assigned strictly on the basis of seniority} (Synchrony) (WSJ\_0286)  \label{ex:mweconn}
\end{exe}

Both Example \ref{ex:conn1} and \ref{ex:mweconn} involve a three word sequence annotated as connectives but in Example \ref{ex:conn1} each token signals a different relation whereas in the latter example they are part of the same connective, hence signal only one relation. Therefore, correct prediction of the boundaries of the connectives is as crucial as identifying them as such because failing to do so may cause whole system to miss existing relations or add artificial ones. To this end, unlike previous studies \cite{hooda2017argument,knaebel2019window}, we assign different labels to single and multi token connectives (\#Conn and \#MWconn respectively) so at the time of inference, we can decide whether consecutive tokens predicted as connective is a part of a multiword connective or signal different relations. 

One drawback of using the publicly available BERT model is that text spans longer than 512 wordpieces cannot be encoded \cite{devlin2018bert}. Therefore, we decided to split the text into paragraphs to ensure the coherence of the text segments as splitting into an arbitrary number of sentences would risk having incoherent segments. Manual inspection of the training data reveals that majority of the relations (84.94\%) have both of their arguments in the same paragraph which further support our decision. 

\subsection{Argument Extraction} \label{sec:arg}

The argument extractor needs to identify the Arg1 and Arg2 spans of each predicted connective. The extractor searches for the arguments of the relation within a window of 100 words centered around the discourse connective \footnote{For multiword connectives, we center the window around the first token.}. Following the IOB2 format, the first word of each argument is tagged as \#ARGX-B while other words within the argument spans are simply labeled as \#ARGX where X is the argument number (1 or 2). The words outside of the relations are labeled as \#NONE.  

The window size is determined by considering the number of relations that can be covered and the label distribution in the extracted spans as longer windows sizes introduce a high of number of \#NONE labels which negatively affect the training. Based on the span lengths of the relations in the training data (Table \ref{tab:span}), window size of 100 presents itself as the best candidate since longer windows only minimally increase the coverage.

\begin{table}[t]
\centering
\begin{tabular}{>{\centering\arraybackslash}p{4cm}c}
 \# of Annotations (\%) & Span Length  \\ \hline
6231 (42.32\%) & $ < 25 $ \\
12243 (83.16\%) & $ < 50 $ \\
13810 (93.81\%) & $ < 75 $ \\
14240 (96.73\%) & $ < 100 $ \\
14617 (99.29\%) & $ < 250 $ \\ \hline
Average & 36.79 \\\hline
\label{tab:span}
\end{tabular}
\caption{Number of annotations with various span lengths, in terms of the number of words in the relation, in the training set which consists of 14722 relations in total.}
\end{table}

\section{Experiments}
Following the CoNLL 2015 setting, we use the PDTB Sections 2-21, 22 and 23 as the training, development and test set respectively. We use the cased BERT$_{base}$ model in our experiments\footnote{https://github.com/google-research/bert} and the classifiers are implemented using the Huggingface's Transformer library\footnote{https://github.com/huggingface/transformers}. The maximum sequence length is set to 400 for connective identification and 250 for argument extraction. We use AdamW optimizer with the learning rate of $5\times 10^{-5}$ and $\epsilon = 10^{-8}$. Both classifiers are fine-tuned for 3 epochs. We train each classifier for 4 runs in order to estimate the variance and report the average performance.

\section{Results and Discussion}

We evaluate our model using the official evaluation script of the CoNLL 2016 shared task\footnote{https://github.com/attapol/conll16st}. The script calculates the exact match scores of the identified connectives, extracted spans of the first and the second argument separately as well as the identification of the both arguments together (Arg1+Arg2).

\begin{table}[]
\centering
\begin{tabular}{p{0.8cm}|ccc|ccc|ccc|ccc}
 & \multicolumn{3}{c}{Conn} & \multicolumn{3}{c}{Arg1} & \multicolumn{3}{c}{Arg2} & \multicolumn{3}{c}{Arg1+Arg2} \\ \hline
 & P& R& F& P& R& F& P& R& F& P& R& F \\ \hline
\cite{wang2015refined}&94.83&93.49&94.16&51.05&50.33&50.68&77.89&76.79&77.33&45.54&44.90&45.22\\ 
\cite{stepanov2015unitn}&-&-&92.77&-&-&50.05&-&-&76.23&-&-&44.58 \\
\cite{yoshida2015hybrid}&91.8 &86.6& 89.1 & 47.5& 44.8& 46.1 &70.5& 66.4 &68.4& 40.0& 37.7& 38.8\\
\hline
\cite{nguyen2016sdp}&83.42&92.22&87.6&51.25&56.65&53.81&68.36&75.57&71.79&43.12&47.66&45.28\\
\cite{qin2016shallow}&92.42&94.88&93.63&49.73&51.06&50.38&75.73&77.75&76.73&44.31&45.49&44.9\\
\cite{li2016constituent}&\textbf{99.67}&\textbf{98.19}&\textbf{98.92}&42.47&41.84&42.15&76.06&74.92&75.48&36.51&35.97&36.24\\
\hline
\cite{knaebel2019window} &71.35& 62.73& 66.76&33.16& 29.15& 31.03&52.47& 46.13& 49.09&37.25& 32.75 &34.86 \\ 
\cite{knaebel2019window}* &-&-& -&46.69 &44.59 &45.62 & 68.94 &65.83 &67.35&48.16 &45.99 &47.05 \\ \hline
Ours&96.62&96.93&96.77&\textbf{60.02}&\textbf{60.22}&\textbf{60.12}&\textbf{80.37}&\textbf{80.63}&\textbf{80.50}&\textbf{53.20}&\textbf{53.37}&\textbf{53.28}\\ \hline
\end{tabular}
\caption{Exact match results (precision, recall, F-score) of explicit discourse relation labeling on PDTB test set. The models within horizontal lines are the best performing systems of CoNLL 2015, CoNLL 2016 and the feature independent systems respectively. *refers to the results when the gold connectives are provided to the model.}
\label{tab:results}
\end{table}

We compare our results with the top performing systems of CoNLL 2015/16 Shared tasks as well as with \cite{knaebel2019window} which is the only feature-independent study that can run on raw texts (Table \ref{tab:results}). We selected the top systems in each sub-task from CoNLL 2016 whereas for CoNLL 2015 we chose the top 3 ranked systems as \cite{wang2015refined} single-handedly achieved the best score in each sub-task that year.

\paragraph{\textbf{Connective Identification:}} In line with the previous work, connective identification is the easiest step where our model achieves almost 97\% F-score. Since the standard deviation among different runs is pretty low ($< 0.5$), we randomly selected one run and manually checked the predicted connectives. In total, 16 unique text spans are incorrectly predicted as discourse connective for a total of 28 times where most of them are \textit{and} tokens (30.6\%). Similarly, \textit{and} also constitutes the 30\% of the false negatives (the connectives which are \textit{not} labeled as such by the classifier), suggesting that \textit{and} is more challenging to disambiguate in term of its discursive role than other connectives. Finally, of all predictions, only two of them  (\textit{10 minutes} and \textit{end}) are not valid connective candidates which further proves the model's success on connective identification. However, since there is not any unseen connective in the test set, we cannot draw any conclusions regarding the generalization capabilities of the proposed model which will be further examined in a future study.

\paragraph{\textbf{Argument Extraction:}} The proposed model achieves the state-of-the-art results in separate extraction of the arguments as well as the full relation extraction.  The increase in the extraction of the first argument is of special importance because the first argument is the most challenging component to automatically predict as it can reside anywhere in the text and do not have any syntactic bounds with the connective, unlike the second argument. 

Manual analysis of the predicted first argument spans reveals that 20\% of all mismatches are only by one or two words and mostly occur in the beginning of the argument. Several cherry-picked  examples are provided in Example \ref{ex:pred} where the predicted spans are underlined and the gold spans are shown in bold.

\begin{exe} \label{ex:pred}
\ex 
\begin{xlist} 
\ex \ul{I expect \textbf{the market to open weaker Monday}}
\ex \ul{crumbled. \textbf{Arbitragers couldn't dump their UAL stock}}
\ex \textbf{This \ul{has both made investors uneasy and the corporations more vulnerable}}
\end{xlist}
\end{exe}

To further investigate the performance of the argument extraction, we ran two additional evaluations. Firstly, we evaluated the performance on the relations where the second argument precedes the first one in the text (e.g. Ex. \ref{ex:alt}) which is quite infrequent (less than 10\% the relations have this structure in the test set). However, the proposed model turned out to be quite successful in those relations and achieves 75.6\% F-score in full relation extraction, suggesting that it learned the argument structure of the discourse connectives considerably well. 

In the second evaluation, we focused on the relations with discontinuous spans where there are at least a five word sequence which do not belong to the any part of the relation between the first argument and the connective. There are 93 such relations in the test set and they are the most challenging ones spreading over a text span of 91 words on average. Unfortunately, the proposed model fails to extract the arguments of those relations by achieving only 14.7\% F-score in the extraction of the full relation, hence extraction of the arguments which are not located in the immediate vicinity of the connective still remains a challenge. Yet, it should also be noted that some of these relations falls outside of the argument extractor's scope due to its window size (see Section \ref{sec:arg}).

\section{Conclusion} 

We have shown that labeling explicit discourse relations is possible without any feature engineering. We achieve state-of-the-art results by finetuning a pre-trained language model on PDTB 2.0 which is the first time that a feature-independent system outperforms the existing knowledge intensive systems on this task. However, detailed evaluations reveal that there is much room for improvement, especially in identifying the discontinuous relations where the arguments are interrupted by various text spans. We see the proposed system as a first step towards a high-performance shallow discourse parser that can be extended to any language with a sufficient annotated data and a pre-trained language model.

\paragraph{\textbf{Acknowledgments}} I would like to thank Robert \"{O}stling and Ahmet \"{U}st\"{u}n for their useful comments and NVIDIA for their GPU grant.

%
%
\bibliographystyle{splncs04}
\bibliography{ref}
\end{document}